\documentclass[lettersize,journal]{IEEEtran}
\usepackage{cite}
\usepackage[numbers]{natbib}
\usepackage{amsmath,amssymb,amsfonts}
\usepackage{algorithmic}
\usepackage{graphicx}
\usepackage{subcaption}
\usepackage{textcomp}
\usepackage{xcolor}

\usepackage{lipsum}
\usepackage{booktabs}
\usepackage{url}

\begin{document}

\title{BOOST-RPF: Boosted Sequential Trees for Radial Power Flow}

\author{Ehimare Okoyomon, Christoph Goebel
\thanks{E. Okoyomon and C. Goebel are with the Professorship of Energy Management Technologies at the Technical University of Munich. }
\thanks{Correspondence to: Ehimare Okoyomon \textless e.okoyomon@tum.edu\textgreater.}
}

\maketitle

\begin{abstract}
Accurate power flow analysis is critical for modern distribution systems, yet classical solvers face scalability issues, and current machine learning models often struggle with generalization. We introduce BOOST-RPF, a novel method that reformulates voltage prediction from a global graph regression task into a sequential path-based learning problem. By decomposing radial networks into root-to-leaf paths, we leverage gradient-boosted decision trees (XGBoost) to model local voltage-drop regularities. We evaluate three architectural variants: Absolute Voltage, Parent Residual, and Physics-Informed Residual. This approach aligns the model architecture with the recursive physics of power flow, ensuring size-agnostic application and superior out-of-distribution robustness. Benchmarked against the Kerber Dorfnetz grid and the ENGAGE suite, BOOST-RPF achieves state-of-the-art results with its Parent Residual variant which consistently outperforms both analytical and neural baselines in standard accuracy and generalization tasks. While global Multi-Layer Perceptrons (MLPs) and Graph Neural Networks (GNNs) often suffer from performance degradation under topological shifts, BOOST-RPF maintains high precision across unseen feeders. Furthermore, the framework displays linear \textit{O(N)} computational scaling and significantly increased sample efficiency through per-edge supervision, offering a scalable and generalizable alternative for real-time distribution system operator (DSO) applications.

\end{abstract}


\section{Introduction}

Accurate and scalable power flow analysis is essential for operating modern distribution networks, especially as distributed energy resources and real-time control applications proliferate. While classical solvers like Newton-Raphson (NR) provide high fidelity, their computational cost and convergence limitations motivate the exploration of fast, data-driven alternatives, particularly in large-scale or real-time settings.

Recent years have seen substantial progress in machine learning-based power flow solvers. Most existing approaches formulate voltage prediction as a \emph{global regression task}, in which the full vector of bus voltages is predicted simultaneously from nodal injections and network parameters. This paradigm has been explored using tabular regression models, physics-informed neural networks, and graph neural networks (GNNs) that exploit the relational structure of the grid through message passing. Several of these models demonstrate impressive in-distribution accuracy and, in some cases, limited generalization across network sizes or topologies. However, large performance degradation under out-of-distribution (OOD) conditions remains a persistent challenge, as highlighted by recent benchmarking efforts \citep{okoyomon2025framework}.

In contrast, classical distribution-level power flow models exploit a fundamentally different inductive bias. Analytical methods such as DistFlow and LinDistFlow \citep{baran1989network} compute voltages \emph{iteratively and locally}, propagating voltage drops from the substation to downstream buses using branch-level equations. These models are inherently size-agnostic, scale linearly with network depth, and rely on the observation that voltage behavior in radial feeders is governed by repeated applications of the same local physical relationships. Despite these advantages, most modern ML solvers abandon this sequential inductive bias in favor of global, graph-level predictions.

This paper revisits the iterative structure of distribution-level power flow for specifically radial distribution grids. We reformulate data-driven power flow from a graph-based regression task to a \emph{sequential path-based learning problem}. Instead of a joint global prediction, we learn \emph{local} voltage-drop mappings between parent and child buses. By decomposing radial networks into ordered root-to-leaf paths, we can leverage established sequence models, such as gradient-boosted decision trees (XGBoost), trained on per-edge supervision across diverse networks.

Our approach is based on the hypothesis that local voltage-drop regularities are largely invariant to global network size and topology. This formulation offers two primary advantages: (i) higher sample efficiency through edge-level training data, and (ii) superior OOD generalization across unseen feeders by aligning the architecture with the recursive physics of power flow. Furthermore, the method supports physics-informed residual learning, where models predict corrections to analytical baselines like LinDistFlow. We benchmark our sequential method against analytical solvers and state-of-the-art neural approaches, using datasets based on the Kerber Dorfnetz and the ENGAGE generalization test suite.

The main contributions of this work are centered around the development of BOOST-RPF, a method that bridges classical distribution-level heuristics with modern sequence modeling:

\begin{itemize}
    \item \textbf{BOOST-RPF Models:} We transform voltage prediction from a global graph problem into a sequential path-based learning task, enabling the use of simple ML models specializing on local physical behavior.
    \item \textbf{Physics-aligned inductive bias:} By training on edge-level voltage drops and residuals of analytical baselines, BOOST-RPF leverages known physics for first-order behavior while learning higher-order nonlinearities from data.
    \item \textbf{Superior Generalization:} We demonstrate that BOOST-RPF is inherently size- and topology-agnostic. The local, compositional mapping mirrors iterative analytical solvers, enabling strong zero-shot generalization across unseen feeders.
\end{itemize}
\section{Literature Review}

\subsection{Analytical Iterative Power Flow Methods}

Classical power flow solvers for radial distribution networks are inherently iterative and local. The DistFlow and LinDistFlow models introduced by Baran and Wu \citep{baran1989network} compute voltage drops recursively along feeder branches, forming the basis for many distribution-level optimization and control methods. Numerous variants refine these models through improved linearizations, multiphase extensions, or optimized coefficients \citep{bernstein2018load, nrel_lindistflow}.

Forward-backward sweep algorithms represent another widely used family of iterative solvers, alternating between current updates from leaves to root and voltage updates from root to leaves until convergence \citep{eminoglu2005new}. More recent approaches rely on successive linear approximations of AC power flow, solving a sequence of linearized subproblems to recover nonlinear voltages \citep{molzahn2019survey}.

While these analytical methods explicitly exploit the ordered structure of radial feeders, they rely on hand-crafted equations and do not leverage data-driven learning to model voltage drops or residual nonlinearities.

Another major family of analytical solvers is based on Newton–Raphson–style methods, which formulate AC power flow as a system of nonlinear algebraic equations and iteratively solve it using local linear approximations. At each iteration, the nonlinear equations are linearized around the current operating point via the Jacobian matrix, and voltage updates are obtained by solving the resulting linear system \citep{dutto_on-extending_2019}. These methods tend to be very accurate but are computationally expensive (due to repeated Jacobian factorizations) and are not guaranteed to converge, particularly in large-scale or heavily-loaded distribution networks.

\subsection{Data-driven Power Flow and Voltage Prediction}

Recent years have seen growing interest in replacing or accelerating conventional power flow solvers using machine learning. Most data-driven approaches formulate voltage or power flow prediction as a tabular or graph-based regression task where all bus states are predicted jointly. Particular success has been found in physics-informed methodologies that leverage domain knowledge to enhance the accuracy or robustness of the learning task \citep{Huang_Wang_2023}.

A substantial body of work formulates power flow approximation using strictly tabular, data-driven regression models. Early and influential studies on data-driven power flow linearization (DPFL) learn linear mappings between nodal injections and voltages or line flows directly from data, demonstrating significantly improved accuracy over classical linearizations while retaining low computational complexity \citep{liu2018dpfl}. Subsequent extensions improve robustness and expressiveness, including lifting-dimension regression that applies nonlinear feature transformations prior to linear fitting \citep{guo2019lifting}, noise-aware variants that explicitly account for measurement uncertainty during training~\citep{liu2020noise}, and piecewise linear power flow models to better capture nonlinear and unbalanced behavior in three-phase distribution systems \citep{chen2022piecewise}. More recent works incorporate stronger physics-inspired structure. For example, \citep{hu2020physics} proposes a physics-guided deep neural network for power flow analysis that augments a fully connected architecture with regularization terms inspired by Kirchhoff’s laws, while \citep{yang2019fast} develops a deep learning surrogate for probabilistic power flow, similarly using branch flow equations as penalty terms into the objective. Overall, while these tabular and physics-informed methods demonstrate that data-driven linearization can outperform analytical approximations, these models have the fundamental limitation that their architectures assume a fixed network size.

Beyond tabular-based learning schemes, Graph neural networks (GNNs) have emerged as a dominant paradigm in this line of work due to the flexibility advantages of message passing. \citep{donon2020graph} proposed the Graph Neural Solver, which learns iterative message passing rules that minimize violations of Kirchhoff’s laws and generalizes across network sizes. Subsequent works extended this ``grid as a network'' approach with improved message passing architectures, such as TAGConv \citep{lin2024powerflownet}, Graph Attention Network \citep{Jeddi_Shafieezadeh_2021}, Recurrent GNN \citep{Böttcher_Wolf_Jung_Lutat_Trageser_Pohl_Ulbig_Grohe_2023} and ARMA GNN \citep{hansen_power_2022}, or line-graph formulations that explicitly model branch interactions \citep{hansen_power_2022}. Furthermore, physics-informed techniques have also been employed to incorporate domain knowledge in the learning task through integration of power balance equations, residuals modeling from a surrogate task such as DC PF, or complex-valued neural networks \citep{okoyomon2025physics, Huang_Wang_2023}. 

Furthermore, several studies benchmark GNN-based solvers against simpler tabular or multilayer perceptron (MLP) baselines. Their results show that the prediction task remains fundamentally global in nature, with MLPs being able to outperform the graph models in static grid configurations \citep{hansen_power_2022, yaniv2025benchmarking}. While GNNs often improve flexibility and performance for unknown grid settings, the generalization problem still largely persists, with model performance dropping by one to two orders of magnitude for unseen grids \citep{okoyomon2025physics, okoyomon2025framework}. The state of the art in data-driven power flow treats voltage prediction as a graph-level learning problem and does not exploit a sequential per-edge voltage-drop formulation that may provide this generalizability.

\subsection{Positioning and Novelty}

To the best of our knowledge, there is no prior work that formulates voltage prediction in distribution networks as a generic ordered parent-child sequence prediction problem and trains generic sequence models such as gradient-boosted trees on per-edge voltage drops and residuals around linearized approximations. Existing ML approaches either (i) operate on the full network graph and predict all bus voltages jointly, or (ii) remain fully analytical while iterating along feeder paths. In contrast, the approach proposed in this paper bridges these paradigms by transforming the voltage prediction task from a graph-based problem into a sequence learning problem, enabling grid-size-independent training, significantly increased sample efficiency through per-edge supervision, and improved generalization grounded in the local physics of voltage drops.
\section{Methodology}

The proposed methodology bridges classical power systems heuristics with state-of-the-art sequential machine learning. We transition from a global network representation to a localized, recursive framework that aligns with the physical topology of radial distribution systems.

\begin{figure}[t!]
    \centering
    \includegraphics[width=0.9\linewidth]{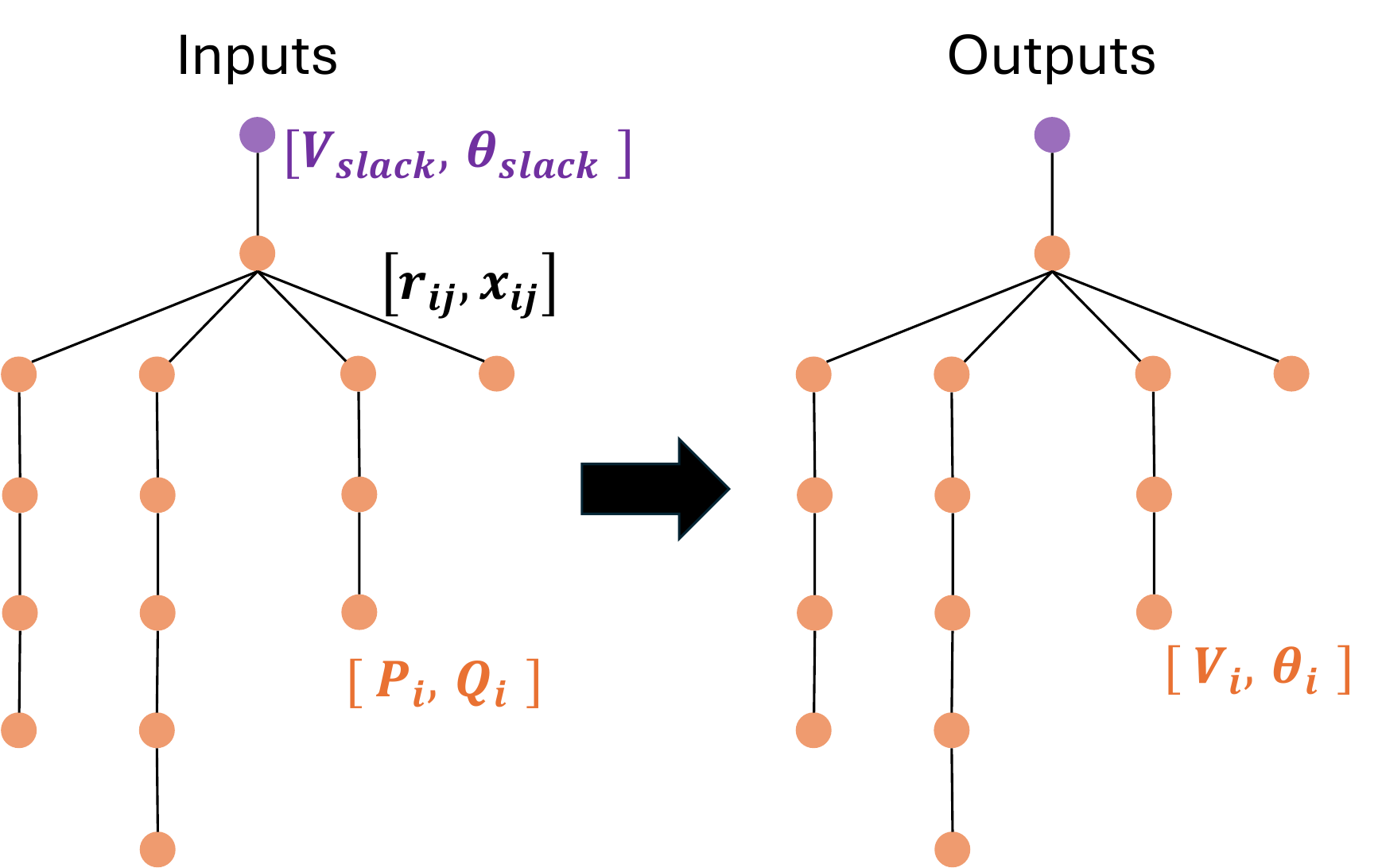}
    \caption{Simplified graphical depiction of the voltage prediction task for PQ (orange) buses.}
    \label{fig:powerflow_overview}
\end{figure}

\subsection{Problem Formulation}

The objective of AC power flow analysis is to determine the active power, reactive power, voltage magnitude, and voltage angle of every bus in a power network, given a set of known and unknown parameters. 
For distribution grids, this typically reduces to voltage prediction: given the slack bus voltage and the complex power quantities at every other bus (PQ bus), the task is to estimate the unknown voltages (Figure \ref{fig:powerflow_overview}). However, the relationship between the power and voltage quantities are governed by complex non-linear functions. Given a set of buses in a network, $\mathcal{N}$, and lines connecting them, the following power injection equations (\ref{eq:powerflow_active_power}–\ref{eq:powerflow_reactive_power}) must be satisfied:

\begin{equation}
\label{eq:powerflow_active_power}
P_{i}
= 
|V_i| \sum_{j=1}^n |V_j| (G_{ij}\cos\theta_{ij}
+
B_{ij}\sin\theta_{ij}),
\qquad
i \in \mathcal{N}
\end{equation}

\begin{equation}
\label{eq:powerflow_reactive_power}
Q_{i}
=
|V_i| \sum_{j=1}^n |V_j| (G_{ij}\sin\theta_{ij}
-
B_{ij}\cos\theta_{ij}),
\qquad
i \in \mathcal{N}
\end{equation}

\noindent Where:
\begin{itemize}
    \item $P_{i}$ is the net active power injection at bus $i$
    \item $Q_{i}$ is the net reactive power injection at bus $i$
    \item $|V_i|$ is voltage magnitude at bus $i$
    \item $\theta_i$ is the voltage phase angle at bus $i$, thus $\theta_{ij} = \theta_{i}-\theta_{j}$
    \item $G_{ij}$ and $B_{ij}$ are the real and imaginary parts of the $ij$-th element of the nodal admittance matrix $Y$.
\end{itemize}

\subsection{Grid-to-Sequence Transformation}

\begin{figure}[t!]
    \centering
    \includegraphics[width=\linewidth]{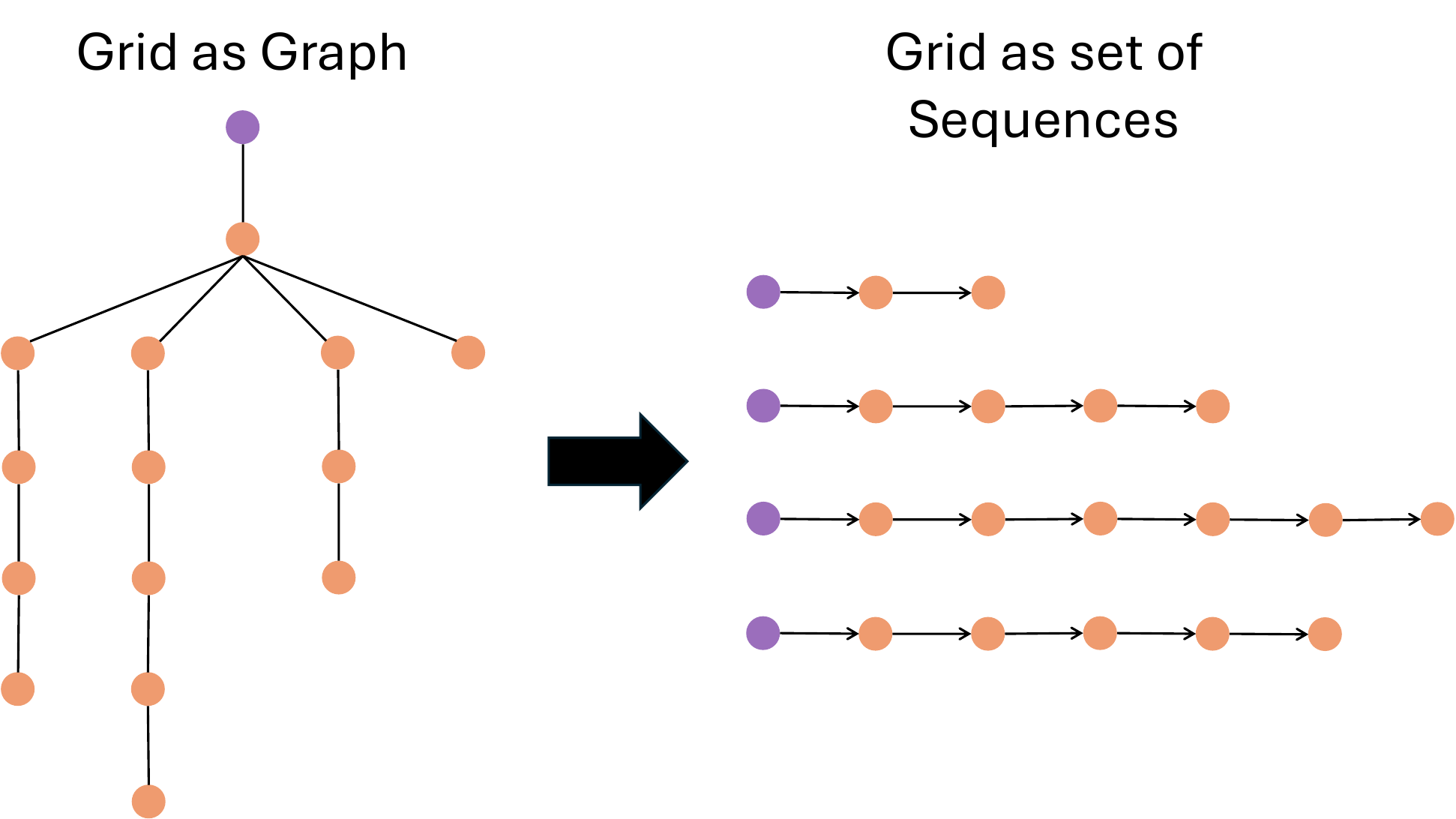}
    \caption{BOOST-RPF task reformulation into sequential learning problem, starting at the slack bus (purple).}
    \label{fig:boost-rpf_overview}
\end{figure}

Traditional AC power flow analysis seeks a global mapping $\mathbf{V} = f(\mathbf{S}, \mathbf{Z})$ for an entire network of $N$ buses. We reformulate this as a \textit{recursive sequential mapping problem} (Figure \ref{fig:boost-rpf_overview}) where the state of a child node $j$ is conditionally dependent on its immediate upstream parent node $i$:
\begin{equation}
V_{j} = f(V_{i}, Z_{ij}, S_{j}, S_{agg, j})
\end{equation}
where $V_j$ is the complex voltage, $Z_{ij} = r_{ij} + jx_{ij}$ is the branch impedance, $S_j = P_j + jQ_j$ is the local power injection, and $S_{agg, j}$ represents the aggregated demand of the downstream sub-tree rooted at $j$. This has the advantage of modeling complete grid predictions as a combination of local node predictions, allowing voltage prediction to flexibly scale to radial architectures of arbitrary size. Additionally, since a single graph can be decomposed into many paths, we are able to have increased sample efficiency when training machine learning models.

\subsubsection{Justification and Decoupling}
By modeling the grid as a Directed Acyclic Graph (DAG) starting from the slack bus, we decompose the complex non-linear system into a series of localized dependencies. We assume that active and reactive power injections ($P, Q$) at individual buses are independent stochastic variables reflecting decoupled consumer behavior. Conversely, bus voltages ($V, \delta$) are treated as strictly dependent variables, physically coupled through the recursive voltage drops along the feeder.

\subsection{Feature Engineering and Pre-processing}

The central conflict in power flow is that voltage propagates downstream but current is determined by upstream accumulation. To address this conflict, we implement a dual-pass pre-processing pipeline, similar to LinDistFlow.

\subsubsection{Backward Power Accumulation}
We perform an $O(N)$ leaf-to-root traversal to calculate the total downstream demand for each node, ignoring line losses for the initial feature set. To maintain consistency with the net injection convention introduced in Equations \ref{eq:powerflow_active_power}–\ref{eq:powerflow_reactive_power} (where loads are represented as negative values), we define the aggregate downstream for a node $j$ with a set of children $\mathcal{C}_j$ as:

\begin{equation}
P_{agg, j} = -P_j + \sum_{k \in \mathcal{C}_j} P_{agg, k}
\end{equation}
\begin{equation}
Q_{agg, j} = -Q_j + \sum_{k \in \mathcal{C}_j} Q_{agg, k}
\end{equation}
These features serve as a proxy for the total current flowing through branch $(i, j)$, enabling the model to account for downstream-induced voltage drops of the entire feeder branch.

\subsubsection{LinDistFlow Baseline}
To provide a physical anchor for the model, we can generate a first-order baseline for voltage magnitude ($V_j$) and voltage angle ($\theta_j$). We choose LinDistFlow because it is an established voltage approximation technique for distribution grids, that follows a similar voltage drop structure in its formulation. Starting from the LinDistFlow simplification of the branch flow equations \citep{baran1989network}:

\begin{equation}
V_j^2 \approx V_i^2 - 2(r_{ij}P_{agg, j} + x_{ij}Q_{agg, j})
\end{equation}
We derive a linear baseline for the voltage magnitude using a first-order Taylor expansion\footnote{The expression $V_j \approx V_i - \frac{r_{ij}P_{agg, j} + x_{ij}Q_{agg, j}}{V_0}$ is obtained by expanding the squared-voltage relationship around a nominal voltage $V_0$ (typically the slack bus) and is utilized here to ensure the machine learning target (magnitude) and the baseline reside in the same linear domain.}:
\begin{equation}
V_{j, \text{LDF}} \approx V_i - \frac{r_{ij}P_{agg, j} + x_{ij}Q_{agg, j}}{V_0}
\end{equation}

To recover the approximate phase angle $\theta_j$ sequentially, we utilize a complementary linearization for convex relaxations of Optimal Power Flow, as detailed by \citep{farivar2013branch}:
\begin{equation}
\theta_{j, \text{LDF}} \approx \theta_i - \frac{x_{ij}P_{agg, j} - r_{ij}Q_{agg, j}}{V_{0}^2}
\end{equation}

\subsubsection{Graph-to-Path Conversion}
The grid topology is finally decomposed into a set of unique root-to-leaf paths $\mathcal{P} = \{p_1, p_2, ..., p_m\}$. Each path $p$ is treated as a sequence of feature vectors \\
$\mathbf{X} = [\mathbf{X}_{root}, ..., \mathbf{X}_{leaf}]$, where the feature vector for node $j$ is constructed as:
\begin{equation}
\mathbf{X}_j = [V_i, \theta_i, r_{ij}, x_{ij}, P_j, Q_j, P_{agg,j}, Q_{agg,j}, V_{j, \text{LDF}}, \theta_{j, \text{LDF}}]
\end{equation}
This sequential structure allows the proposed models to maintain a continuous ``memory'' of the cumulative voltage drop and angle shift from the slack bus to the feeder extremities.

\subsection{Sequential Learning Architectures} \label{method:sequential-learning-architecture}

To predict voltages, we leverage the well-established XGBoost model \citep{chen2016xgboost} for sequential learning. XGBoost utilizes gradient-boosted decision trees to map an input feature vector $\mathbf{X}_i$ to a target output through an iterative, additive process. Rather than training trees in parallel, the algorithm initializes a base prediction, $f_0$, and subsequently trains each new decision tree, $f_t$, to predict the residuals (i.e. the gradient of the loss function) from the previous predictor, $f_{t-1}$. This process continues for a fixed number of iterations, progressively minimizing the objective function.

For our sequential model, we create three XGBoost variants for testing, each with a modified learning target:

\begin{itemize}
    \item \textbf{Variant A: Absolute Voltage.} The model predicts the voltage quantities $V_{j, \text{true}}$ directly, using the feature vector $X_i$:
    \begin{equation}
    \text{target} = V_{j, \text{true}}
    \end{equation}
    \item \textbf{Variant B: Parent Residual (Voltage Drop).} The model predicts the direct drop from the parent state:
    \begin{equation}
    \text{target} = \epsilon_{drop} = V_i - V_{j, \text{true}}
    \end{equation}
    \item \textbf{Variant C: Physics-Informed Residual.} The model predicts the deviation from the linear approximation:
    \begin{equation}
    \text{target} = \epsilon_{physics} = V_{j, \text{LDF}} - V_{j, \text{true}}
    \end{equation}
    This target leverages known physics, requiring the model to learn only the high-order nonlinearities and line losses.
\end{itemize}

\subsection{Implementation Strategy: BFS-Based Inference}

Inference is executed via a Breadth-First Search (BFS) traversal starting at the slack bus,
ensuring that we only visit each node a fixed number of times along the chain, thus enabling efficient scaling.

\textbf{High-Level Algorithm:}
\begin{enumerate}
    \item \textit{Path Decomposition:} Use BFS to traverse the graph and determine paths (root-to-leaf sequences) based on parent-child connections and visiting order.
    \item \textit{Backward Pass:} Aggregate $P, Q$ demands from leaves to root using parents.
    \item \textit{Sequential Prediction:} Perform ML inference along paths, in visiting order, keeping track of the target node. The output of every previous node can thus be used as an input for a future prediction.
    \item \textit{Consensus:} Concatenate predictions into a single vector.
\end{enumerate}

Another obvious scheme would involve solving each path separately in an autoregressive manner, without waiting for parent nodes, and then aggregating overlapping node predictions from different paths to resolve voltage. However, this approach entails an $O(N^2)$ computational cost due to redundant calculations of overlapping sub-paths. While such schemes can be easily made parallel and can achieve $O(\log N)$ latency on balanced trees with sufficient hardware, they remain computationally wasteful compared to our proposed method. By contrast, our chaining algorithm achieves $O(N)$ total complexity and can similarly be parallelized by tree depth to reach $O(\log N)$ latency without the overhead of redundant predictions.

\begin{table}[t]
\centering
\caption{Experiment Data}
\label{tab:experiment-data}
\begin{tabular}{lccc}
\toprule
\textbf{Grid} & \textbf{Samples} & \textbf{Buses} & \textbf{Branches} \\
\midrule
\textbf{LV Rural1} & 300 & 15 & 14 \\
\textbf{LV Rural2} & 300 & 97 & 96 \\
\textbf{LV Rural3} & 300 & 129 & 128 \\
\textbf{LV Semi-urban4} & 300 & 44 & 43 \\
\textbf{LV Semi-urban5} & 300 & 111 & 110 \\
\textbf{LV Urban6} & 300 & 59 & 58 \\
\textbf{Kerber Dorfnetz} & 1800 & 116 & 115 \\
\bottomrule
\end{tabular}
\end{table}

\subsection{Evaluation Framework}

To evaluate our methods, we propose three experiment sets, designed to assess an increasingly difficult level of model performance. Our experiments aim to answer the following questions:

\begin{enumerate}
    \item Experiment 1: How well can our model learn voltage prediction of a single grid?
    \item Experiment 2: How well can our model learn voltage prediction from a diverse set of networks?
    \item Experiment 3: How well can our model extrapolate to new voltage prediction contexts?
\end{enumerate}

\subsubsection{Fixed Grid}
For Experiment 1, we use the established Kerber Dorfnetz distribution grid. This is a representative 0.4 kV village distribution grid model characterized by a radial topology, with 116 buses (57 of which are residential load buses) and 115 branches, designed to evaluate the hosting capacity and voltage stability of typical European low-voltage networks under increasing photovoltaic penetration \citep{kerber2011aufnahmefahigkeit}. As standard low-voltage benchmarks are typically provided as passive topological snapshots to allow for user-defined study cases, we must define distributed energy resource (DER) penetrations. To align with near-future scenarios, we aim to match the penetration levels of the SimBench Scenario 1 (future grid with normal DER increase) \citep{meinecke2020simbench}. Thus, the Kerber Dorfnetz was augmented with a stochastic distribution of Photovoltaics (40\% penetration, 5–15 kWp), Electric Vehicles (20\%, 11 kW), Heat Pumps (15\%, 3–6 kW), and Battery Storage (30\% of PV sites, 5 kW/10 kWh). Since each load is constant in the default network, we also initialize base loads by assigning varying residential scales (0.5x, 1.0x, or 2.5x of nominal values) to capture the socio-economic diversity of real-world consumption. By combining these diversified base loads with modern multi-energy assets, we ensure that our base network aligns with our targeted high-variance, future-grid environment.

We employ Powerdata-gen \citep{powerdata-gen} to generate 1800 new training samples from this base network, varying in loading and generation properties. This provides a robust environment for training machine learning models to predict non-linear voltage sensitivities in decentralized distribution systems. By training and testing our methods on this fixed-network dataset, we evaluate if our model can learn the voltage prediction task at all.

\subsubsection{Heterogeneous Grids and OOD Generalization}
The next experiments test model generalization, using the open-source ENGAGE dataset \citep{okoyomon_2025_15464235}. The ENGAGE dataset is a heterogeneous distribution grid dataset, consisting of 10 unique base grids, spanning the low- and medium-voltage levels, with varying network characteristics such as number of nodes, grid types (rural, semi-urban, urban, commercial), line loadings, etc. From this data suite, we use the six LV radial networks, each with their 300 samples, to form our generalization dataset. In our first generalization test (Experiment 2), we mix all the networks together to train our model, forcing it to learn from a heterogeneous set of networks, and verifying that it can be used for a mixture of grids concurrently. In the second generalization test, we perform an out-of-distribution (OOD) generalization evaluation, in which we train our model on the samples from five of the networks, and test on the remaining network samples, which our model has never seen before. This test represents the target goal of power flow modeling: to develop a method that can extrapolate and ultimately be useful and reliable, even in somewhat uncertain situations.

The experiment distribution grids are summarized in Table \ref{tab:experiment-data}.

\subsubsection{Training procedure}
For all experiments, we keep the model pipeline constant. Namely, we use a 4:1:1 split for training, validation, and testing data. This results in 1500 networks used to optimize and fit our model, and 300 remaining networks for testing. Additionally, we hyperparameter tune our models to Experiment 2, to maximize model expressivity while trying to minimize overfitting. We then use the same XGBoost model structure for all further experiments: 200 estimators with max depth of 7 and subsample of 0.9 for regularization. Each model variant (XGB-Absolute, XGB-Parent, XGB-LDF) is thus identical in construction, differing only in prediction target, as explained in Section \ref{method:sequential-learning-architecture}. 

\subsubsection{Metrics}

We quantify performance using the Root Mean Square Error (RMSE) for magnitude ($|V|$) and angle ($\delta$). As additional metrics, we evaluate model training time, inference time scaling, and model size.

\subsection{Model Baselines}

To contextualize the performance of our sequential path-based learning approach, we benchmark it against three distinct classes of models: an analytical approximation, a state-of-the-art Graph Neural Network (GNN), and a global Multi-Layer Perceptron (MLP). This selection ensures a comprehensive comparison across physics-based, graph-aware, and purely data-driven architectures.

\subsubsection{Analytical Approximation}
As a physics-based baseline, we utilize the full \textit{DistFlow} model \cite{baran1989network}, specifically designed for radial distribution systems, whose formulation includes the nonlinear effects of the power lines on the demand aggregation and the voltage drop calculations. This serves as the primary benchmark for simplified modeling accuracy and its linearized version forms the basis for our physics-informed residual learning variant (XGB-LDF).

\subsubsection{Neural Network Baselines}
We implement the neural architectures proposed by \citet{hansen_power_2022} which were originally designed for decentralized power flow tasks. We started with their open-sourced configurations and performed targeted hyperparameter tuning to ensure these baselines reached their peak performance on our specific datasets.

\begin{itemize}
    \item \textbf{ARMA-GNN:} This model leverages Auto-Regressive Moving Average (ARMA) filters to capture graph signals. Unlike standard first-order GCNs, the ARMA layers are designed to better handle long-range dependencies by aggregating information from multiple neighborhoods. The architecture comprises two dense pre-processing layers, followed by eight ARMA convolution layers (with five stacks each), and two post-processing layers. All hidden layers have dimension 64 and utilize LeakyReLU activations ($\alpha = 0.2$).
    \item \textbf{Global MLP:} This model serves as a purely data-driven, topology-agnostic baseline. It processes bus and edge features through independent pre-processing layers (each of dimension 128) before concatenating them for global processing through two wide dense layers (dimension 256). Thus, this model has a complex global view of each network, which it uses to make predictions. Since the model structure is static and topology-agnostic, evaluating several distribution grid types with one MLP requires the network to be built to accommodate the largest network in our dataset.
\end{itemize}

In our preliminary studies, we evaluated several other graph architectures, including PowerFlowNet \citep{lin2024powerflownet}, PowerGraph models \citep{varbella2024powergraph}, and the degree-augmented GCN proposed in the original ENGAGE framework paper \citep{okoyomon2025framework}. However, the ARMA-GNN consistently outperformed these alternatives across our test suites; consequently, it was selected as the representative GNN baseline for this study.

\subsubsection{Tuning and Implementation}

To ensure a fair comparison, we did not simply use "out-of-the-box" settings. All neural models underwent hyperparameter tuning to optimize learning rates, number of layers, and regularization parameters. The models were trained until convergence, signaled by an early stopping based on the validation loss (with a patience of 500 epochs), or until a maximum of 10000 epochs. This ensures that the performance gains observed in our proposed method are due to the architectural shift to path-based learning rather than poorly tuned baselines.

\section{Results}

All models were trained and evaluated on a workstation equipped with an AMD Ryzen 9 5950X (16-core, 32-thread) processor and an NVIDIA RTX 5000 Ada GPU with 32 GB of GDDR6 VRAM, utilizing CUDA 13.0. To ensure reproducibility and account for stochastic variations in model initialization and training, we report the mean and standard deviation across three independent random seeds. 

\subsection{Controlled vs. Heterogeneous Environments}

\begin{table}[t!]
\centering
\caption{Experiment 1 Results. Performance comparison on 300 samples of the Kerber Dorfnetz grid.}
\label{tab:experiment-1-kerber}
\begin{tabular}{lcc}
\toprule
\textbf{Model} & \multicolumn{2}{c}{\textbf{RMSE}} \\
 & \textbf{VM (p.u.)} & \textbf{VA (deg)} \\
\midrule
\textbf{GlobalMLP} & 6.80e-03 {\scriptsize $\pm$ 2.43e-04} & 1.82e-01 {\scriptsize $\pm$ 3.28e-03} \\
\textbf{ARMA-GNN} & 1.05e-02 {\scriptsize $\pm$ 1.04e-04} & 3.78e-01 {\scriptsize $\pm$ 1.81e-01} \\
\textbf{XGB-Absolute} & 4.13e-03 {\scriptsize $\pm$ 1.22e-04} & 2.07e-02 {\scriptsize $\pm$ 1.26e-03} \\
\textbf{XGB-Parent} & \textbf{8.42e-04 {\scriptsize $\pm$ 3.31e-05}} & \textbf{7.34e-03 {\scriptsize $\pm$ 2.50e-04}} \\
\textbf{XGB-LDF} & 2.53e-03 {\scriptsize $\pm$ 4.38e-05} & 1.12e-02 {\scriptsize $\pm$ 9.40e-05} \\
\textbf{DistFlow} & 3.42e-02 {\scriptsize $\pm$ 3.90e-04} & 4.40e-01 {\scriptsize $\pm$ 4.36e-04} \\
\bottomrule
\end{tabular}
\end{table}

\begin{table}[t!]
\centering
\caption{Experiment 2 Results. Performance across a heterogeneous set of LV networks from the ENGAGE dataset.}
\label{tab:experiment-2-all}
\begin{tabular}{lcc}
\toprule
\textbf{Model} & \multicolumn{2}{c}{\textbf{RMSE}} \\
 & \textbf{VM (p.u.)} & \textbf{VA (deg)} \\
\midrule
\textbf{GlobalMLP} & 5.59e-03 {\scriptsize $\pm$ 1.79e-04} & 1.45e-01 {\scriptsize $\pm$ 5.19e-03} \\
\textbf{ARMA-GNN} & 4.21e-02 {\scriptsize $\pm$ 9.04e-03} & 3.66e-01 {\scriptsize $\pm$ 2.70e-01} \\
\textbf{XGB-Absolute} & 4.89e-03 {\scriptsize $\pm$ 1.13e-04} & 2.67e-02 {\scriptsize $\pm$ 2.29e-03} \\
\textbf{XGB-Parent} & \textbf{7.30e-04 {\scriptsize $\pm$ 4.72e-05}} & \textbf{1.12e-02 {\scriptsize $\pm$ 2.71e-03}} \\
\textbf{XGB-LDF} & 5.18e-03 {\scriptsize $\pm$ 2.23e-04} & 4.78e-02 {\scriptsize $\pm$ 3.01e-03} \\
\textbf{DistFlow} & 2.83e-02 {\scriptsize $\pm$ 1.24e-04} & 1.10e+00 {\scriptsize $\pm$ 1.22e-01} \\
\bottomrule
\end{tabular}
\end{table}

In Experiment 1 (Fixed Grid), we evaluate the models' ability to learn the power flow of a single, well-defined network. As shown in Table \ref{tab:experiment-1-kerber}, the XGB-Parent variant achieves the highest precision, with a Voltage Magnitude (VM) RMSE of $8.42 \times 10^{-4}$ p.u. and a Voltage Angle (VA) RMSE of $7.34 \times 10^{-3}$ degrees, significantly outperforming all other baselines. The neural network baselines, while respectable, show an order of magnitude higher error compared to the sequential XGBoost variants. Notably, the physics-based DistFlow performs the worst in this controlled setting, with an RMSE nearly two orders of magnitude higher than the top-performing model.

Moving to Experiment 2 (Heterogeneous Grids), where models are trained on a mixture of six distinct networks, the performance hierarchy remains stable. Table \ref{tab:experiment-2-all} shows that XGB-Parent maintains its superiority, improving its VM RMSE slightly to $7.30 \times 10^{-4}$ p.u.. Interestingly, the ARMA-GNN's performance degrades significantly when exposed to the heterogeneous dataset, with its VM error increasing five-fold compared to Experiment 1. In contrast, the GlobalMLP and XGB-Absolute variants demonstrate relative stability, suggesting a higher capacity for handling diverse topological features when trained on a mixed distribution.

\begin{table*}[t]
\centering
\caption{Experiment 3 Results. Statistical Distribution of Out-of-Distribution (OOD) Generalization.}
\label{tab:experiment-3-ood}
\begin{tabular}{lcccccccc}
\toprule
\textbf{Model} & \multicolumn{4}{c}{\textbf{RMSE VM (p.u.)}} & \multicolumn{4}{c}{\textbf{RMSE VA (degree)}} \\
 & \textbf{Min} & \textbf{Max} & \textbf{Mean} & \textbf{Std} & \textbf{Min} & \textbf{Max} & \textbf{Mean} & \textbf{Std} \\
\midrule
\textbf{GlobalMLP} & 0.6626 & 2.6433 & 1.4872 & 0.6089 & 41.9360 & 96.6886 & 64.5122 & 16.5791 \\
\textbf{ARMA-GNN} & 0.0089 & 0.0999 & 0.0292 & 0.0229 & 0.3321 & 15.2312 & 2.0956 & 3.4924 \\
\textbf{XGB-Absolute} & 0.0108 & 0.0306 & 0.0200 & 0.0053 & 0.1183 & 1.9460 & \textbf{0.6435} & \textbf{0.5841} \\
\textbf{XGB-Parent} & \textbf{0.0064} & \textbf{0.0215} & \textbf{0.0134} & 0.0050 & 0.0497 & \textbf{1.6498} & 0.7677 & 0.7149 \\
\textbf{XGB-LDF} & 0.0096 & 0.0245 & 0.0187 & \textbf{0.0041} & 0.1518 & 3.4437 & 0.9725 & 1.1056 \\
\textbf{DistFlow} & 0.0140 & 0.0363 & 0.0276 & 0.0073 & \textbf{0.0426} & 4.3472 & 1.0844 & 1.5539 \\
\bottomrule
\end{tabular}
\end{table*}

\begin{figure}[b]
    \centering
    \includegraphics[width=\linewidth]{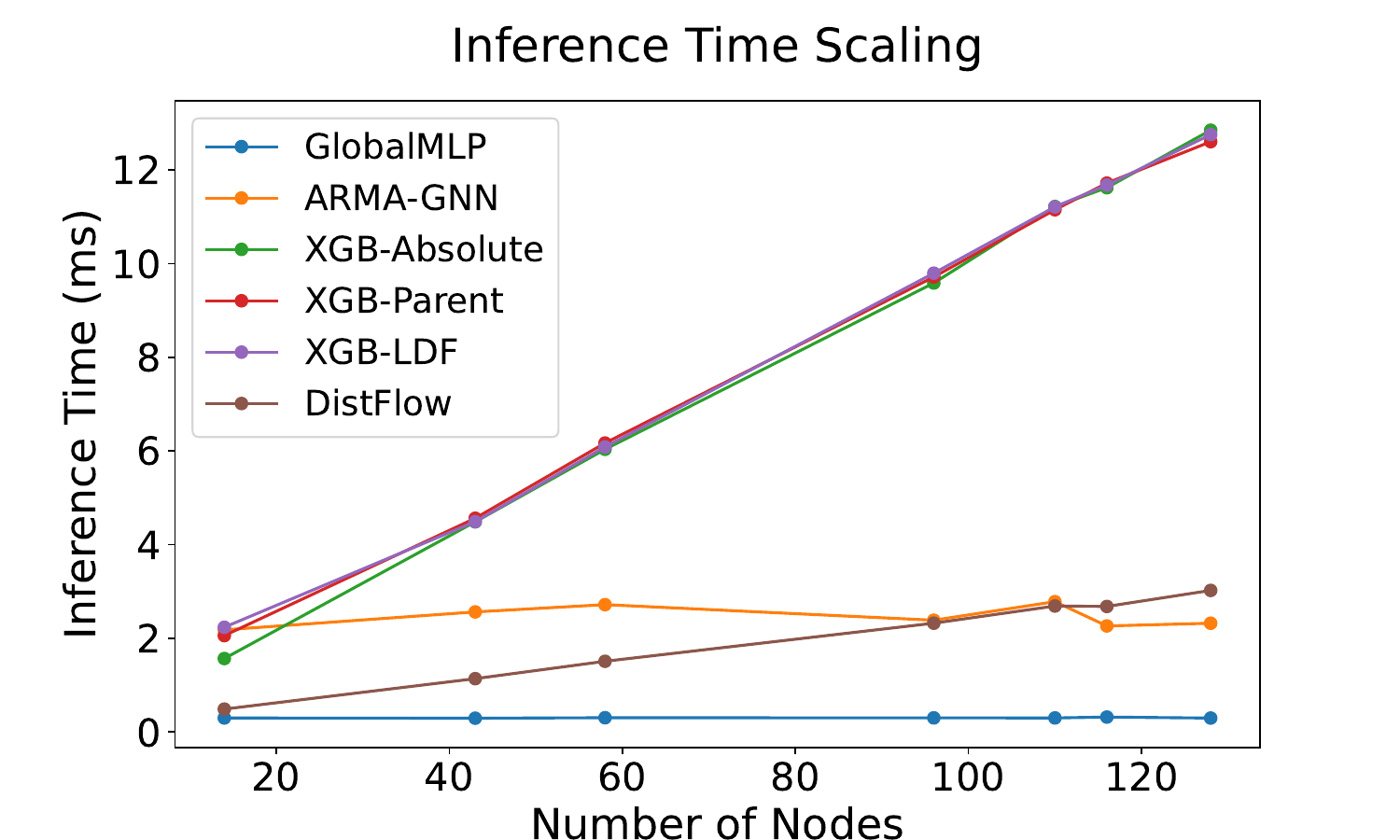}
    \caption{Computational scalability analysis. Average inference time (ms) displayed as a function of the number of nodes in the network.}
    \label{fig:inference_time-scaling}
\end{figure}

\subsection{Out-of-Distribution Generalization}

\begin{figure*}[t]
  \centering
  \begin{subfigure}{0.45\linewidth}
      \centering
      \includegraphics[width=\linewidth]{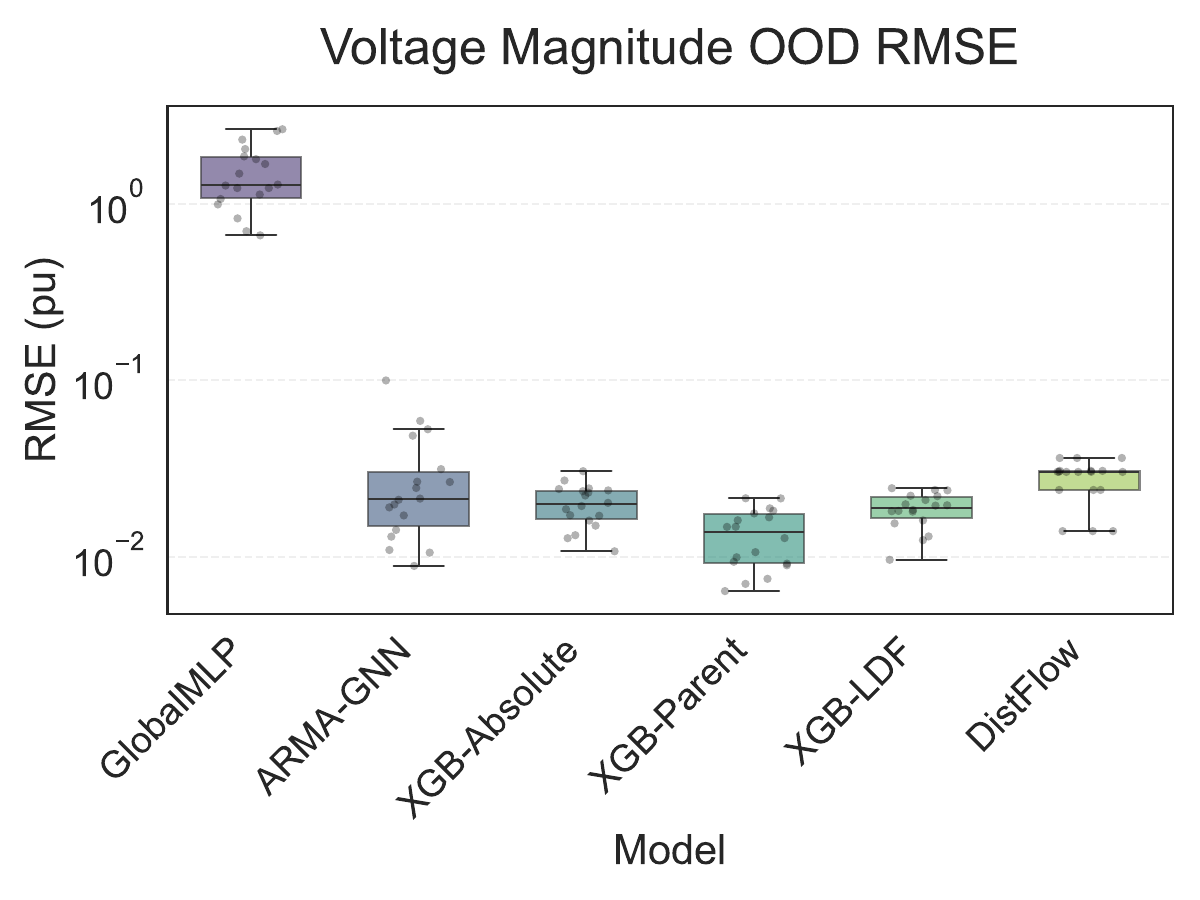}
      \caption{Voltage Magnitude}
  \end{subfigure}
  \begin{subfigure}{0.45\linewidth}
      \centering
      \includegraphics[width=\linewidth]{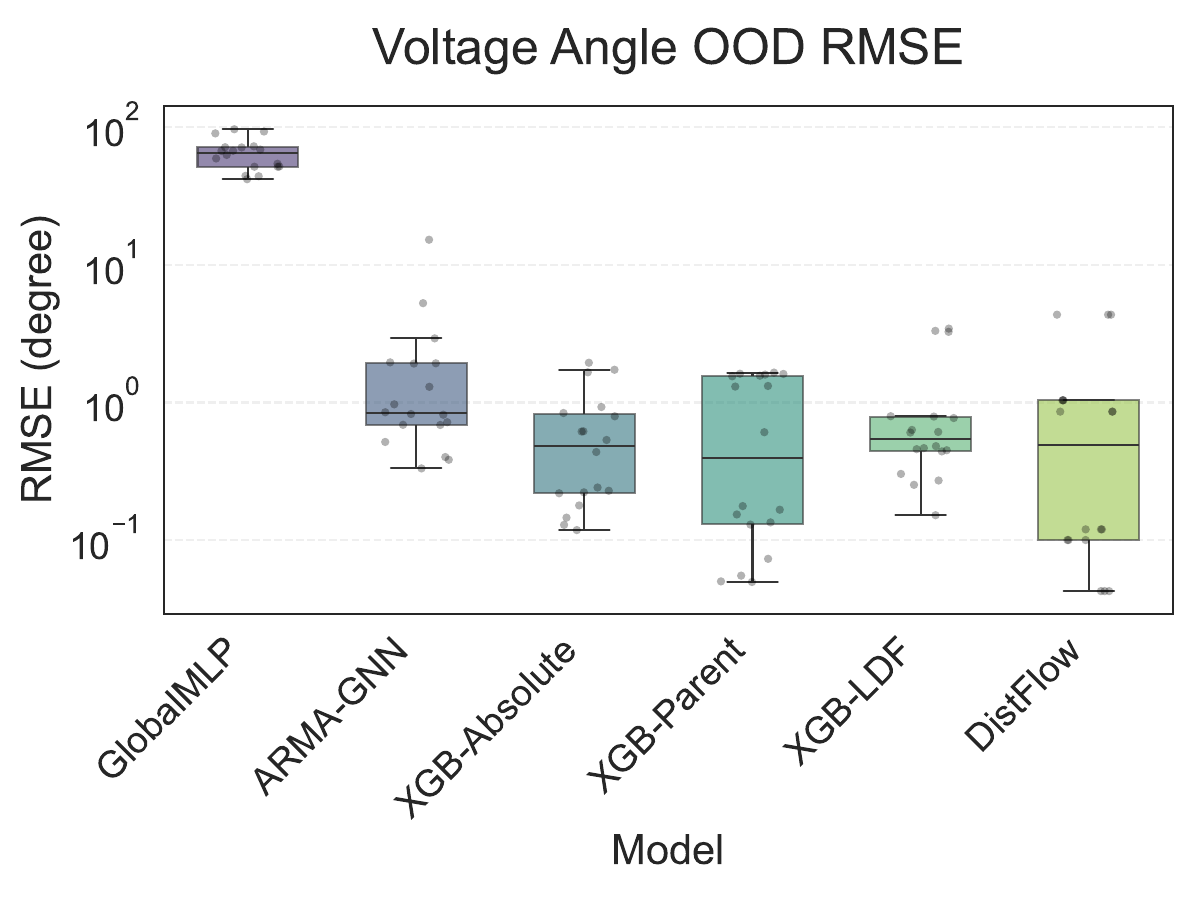}
      \caption{Voltage Angle}
  \end{subfigure} 
  \caption{Error distribution for the out-of-distribution assessment (Experiment 3).}
  \label{fig:ood_rmse}
\end{figure*}

The results for Experiment 3 (OOD), summarized in Table \ref{tab:experiment-3-ood} and Figure \ref{fig:ood_rmse}, highlight critical differences in model robustness when encountering unseen network structures.

\subsubsection{\textbf{GlobalMLP Failure}} The GlobalMLP fails to generalize, exhibiting an average VM RMSE of 1.4872 p.u. and VA RMSE of 64.5122, which is several orders of magnitude worse than its performance in the first two experiments. These predictions render the model unusable for any practical power systems use cases with an unseen network.

\subsubsection{\textbf{Sequential Robustness}} Both XGB-Parent and XGB-Absolute models remain highly competitive. XGB-Parent achieves the best mean VM RMSE (0.0134 p.u.), while XGB-Absolute provides the most accurate mean angle predictions (0.6435 degrees).

\subsubsection{\textbf{Baseline Convergence}} In this OOD context, due to the deterioration of the other models' performances, the analytical DistFlow becomes competitive again with the neural methods, even providing the lowest minimum angle error (0.0426 degrees). The ARMA-GNN also shows relatively strong generalization in terms of voltage magnitude, maintaining a respectable mean VM RMSE of 0.0292 p.u..

\subsection{Computational and Operational Efficiency}

The efficiency metrics from Table \ref{tab:model-efficiency} and Figure \ref{fig:inference_time-scaling} reveal distinct trade-offs between training effort and inference characteristics.

\subsubsection{\textbf{Training Speed}} All XGBoost variants train in approximately 9 to 14 seconds, representing a fraction of the time required for the GlobalMLP (216.12s on average) and especially the ARMA-GNN (1420.77s on average). This is particularly noteworthy as the neural networks utilized GPU acceleration with batching, whereas the sequential XGB models were trained linearly on the CPU.

\subsubsection{\textbf{Inference Scaling}} As shown in Figure \ref{fig:inference_time-scaling}, the GlobalMLP and ARMA-GNN provide constant inference times regardless of network size ($<3$ ms). In contrast, the XGB variants exhibit a clear linear $O(N)$ scaling, with inference time rising to approximately 14 ms for 130 nodes.

\subsubsection{\textbf{Model Size}} Despite the higher parameter count of neural models ($\approx 345\text{K}$ vs $\approx 90\text{K}$ for XGB), all models occupy a similar disk space of approximately 1.4 MB, ensuring a fair comparison for edge-deployment scenarios. Note that ``parameters'' refers to weights and biases in neural models and decision nodes (such as split nodes and leaf values) in XGBoost. As such, each parameter type carries a different computational and memory overhead per unit.

\section{Discussion}

\begin{table}[b]
\centering
\caption{Model Efficiency Metrics. Comparison of training time, total trainable parameters, and final model storage size.}
\label{tab:model-efficiency}
\begin{tabular}{lccc}
\toprule
\textbf{Model} & \textbf{Train time (s)} & \textbf{Params ($\times 10^3$)} & \textbf{Size (MB)} \\
\midrule
\textbf{GlobalMLP} & $216.12 \pm 67.3$ & 346.8K & 1.360 \\
\textbf{ARMA-GNN} & $1420.77 \pm 516.37$ & 343.2K & 1.348 \\
\textbf{XGB-Absolute} & $11.29 \pm 1.66$ & 86.0K & 1.384 \\
\textbf{XGB-Parent} & $11.78 \pm 1.48$ & 86.3K & 1.388 \\
\textbf{XGB-LDF} & $11.77 \pm 1.4$ & 90.3K & 1.448 \\
\textbf{DistFlow} & - & - & - \\
\bottomrule
\end{tabular}
\end{table}

\subsection{Sequence Modeling and Learning Dynamics}

The consistent superiority of the XGBoost models validates the shift from global, graph-level modeling to a sequence learning paradigm. This transformation allows for grid-size-independent training, which ensures the model remains effective regardless of the network's scale. A key driver of this performance is the significantly increased sample efficiency achieved through training on paths. Instead of treating an entire grid as a single training instance, our approach extracts sequences from every root to node traversal, effectively multiplying the training data available from each network sample. For example, from the 1200 training and 300 validation network samples, we are able to obtain 138000 and 34500 paths to learn from, respectively.

The results also show that learning local voltage drop residuals is a far more ``learnable'' task for gradient-boosted trees than predicting absolute voltages or deviations from linearized physics, as evidenced by the consistently superior performance of the XGB-Parent model. This is likely due to a smaller optimization space associated with residual learning, as compared to absolute voltage prediction, and the lack of precision in the linear anchor when applied to distribution networks with high R/X ratios and bi-directional power flows. Evidently, the XGBoost models possess a high capacity to learn. Despite having nearly four times fewer raw parameters than the ARMA-GNN, the model occupies a nearly identical memory footprint. 
This parity suggests that XGBoost parameters provide higher informational utility per unit of storage than individual neural weights. Consequently, the model achieves superior representative capacity and higher accuracy with a much leaner parameterization and a fraction of the training time.

\subsection{Generalization and Model Robustness}

The catastrophic failure of the GlobalMLP in Experiment 3 confirms that topology-agnostic, purely data-driven models are unsuitable for generalizable power flow tasks because they tend to memorize specific network configurations rather than learning underlying physical relationships. In contrast, the ARMA-GNN demonstrates strong generalization, though its high training time may be a bottleneck in environments with frequent distribution drifts or network reconfigurations.

The competitive performance of DistFlow in OOD scenarios is a crucial baseline. When the grid structure is completely unknown, the fact that DistFlow performs similarly to neural methods suggests there is little incentive to deploy complex neural models for entirely new grids unless they can demonstrably surpass this linear approximation. However, our sequential XGBoost models manage to bridge this gap, offering the speed of machine learning with the robustness of a sequential formulation. Considering its fast training time, it would also be reasonable to frequently train on new, live data, quickly improving its performance in new scenarios.

\begin{figure}[b!]
    \centering
    \includegraphics[width=\linewidth]{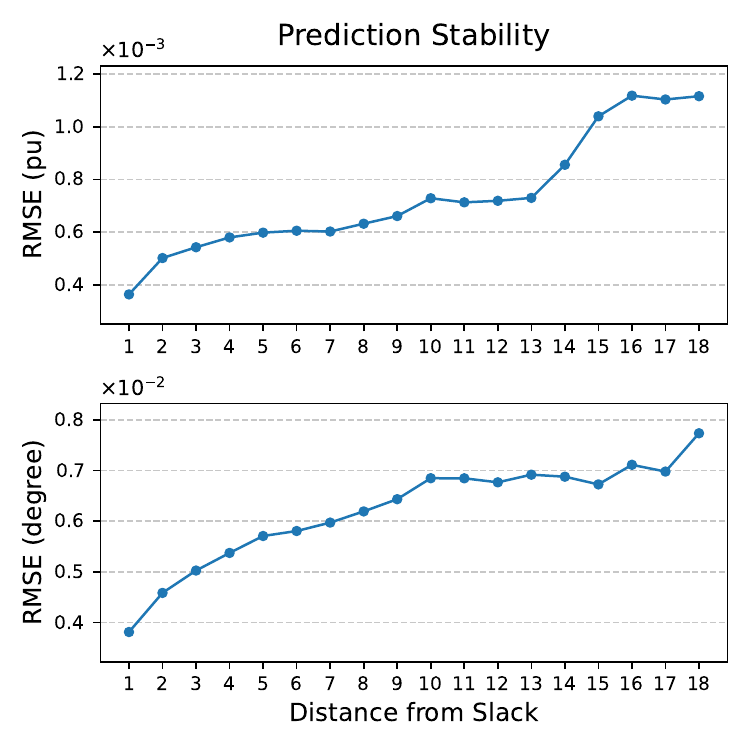}
    \caption{Analysis of XGB-Parent prediction stability across feeder depth of the Kerber Dorfnetz.}
    \label{fig:error_accumulation}
\end{figure}

However, a common concern with sequential architectures is the potential for error accumulation along long feeder paths. To quantify this, we analyzed the RMSE of the XGB-Parent model on the Kerber Dorfnetz dataset as a function of the topological distance (number of hops) from the slack bus (Figure \ref{fig:error_accumulation}). While some accumulation is present, the marginal drift is remarkably low: the total increase in RMSE from the first hop to the last hop $0.0008$ p.u. for voltage magnitude and $0.005^{\circ}$ for voltage angle. The fact that the error remains bounded at such low magnitudes, even at feeder extremities, demonstrates the robustness of the sequential approach for deep radial networks and explains why it maintains competitive accuracy against global, non-recursive models.

\subsection{Scalability and Real-Time Deployment}

The inference results highlight a fundamental trade-off. While the GlobalMLP provides the fastest inference for every grid due to its global, single-step prediction, its lack of generalization is a severe limitation. The slope of the $O(N)$ scaling of the XGBoost models is a result of their Breadth-First Search (BFS) based sequential formulation, where each step requires a return to the Python layer for data transfer to prepare the next autoregressive vector. The demonstrated scaling can thus be viewed as an upper bound on real-world performance. In a production environment, implementing the pipeline in a lower-level language (e.g., C++) would significantly reduce the linear slope by eliminating the Python-to-C overhead. Furthermore, these sequential models can be parallelized by tree depth to reach $O(\log N)$ latency, further outperforming traditional iterative methods. For comparison, the traditional Newton-Raphson AC power flow method requires solving a system of nonlinear equations by building and solving the Jacobian matrix at each iteration. When treated with dense linear algebra, the computational cost scales at $O(N^3)$ with the number of buses, N, due to the cubic cost of solving linear systems \citep{costilla2020combining}. Since realistic grid topologies have sparse Jacobian matrices, efficient factorization methods and sparse network techniques can be used to reduce scaling behaviour closer to $O(N^{1.3} - N^{1.6})$ during runtime \citep{johnson2008sparse}. Thus, for networks exceeding a few hundred nodes, the sequential learning approach is expected to provide a definitive speed advantage over analytical iterative solvers while maintaining the flexibility of a data-driven model. This reduction in latency is critical for high-frequency applications such as real-time hosting capacity analysis or Monte Carlo-based contingency screening, where thousands of power flow scenarios must be evaluated within tight operational windows to ensure grid stability.

\subsection{Model Transparency}

Beyond predictive accuracy, the BOOST-RPF method offers an ancillary benefit in interpretability compared to global neural baselines like the GlobalMLP or ARMA-GNN. By predicting local voltage drops instead of abstract global quantities, the learning target remains physically intuitive and avoids the ``black-box'' feature mixing typical of deep neural architectures. Furthermore, tree-based ensembles are inherently more interpretable than neural networks, allowing for more straightforward auditing of the prediction logic, such as evaluating the impact of specific local branch features, without the need for the complex post-hoc analysis required by deep-learning models.

\section{Limitations and Future Work}

Despite the merits of BOOST-RPF's improvements in generalization and sample efficiency, several limitations should be considered:

\begin{itemize}
\item \textbf{Topological Restriction:} The current BFS-based sequence formulation is strictly optimized for radial (tree-structured) networks, which are common in low-voltage distribution but exclude meshed or looped configurations.


\item \textbf{Python Layer Overhead:} The observed linear $O(N)$ scaling is partially attributed to the frequent data-transfer between the Python layer and the model engine required for the autoregressive preparation step.

\item \textbf{Balanced System Modeling:} The current evaluation assumes a single-phase equivalent model, which may not capture the complex phase-coupling and current non-linearities found in unbalanced three-phase distribution grids.
\end{itemize}

Considering the results of the study, the limitations outlined above, and our specific choices of models and implementation, we see several avenues for future research:  

\begin{itemize}
    \item \textbf{Meshed Grid Adaptation:} Future iterations could investigate extending the sequential paradigm to meshed topologies by turning them into radial equivalents by identifying cycle-breaking nodes or injecting virtual compensation currents at loop-breaking nodes. Early experimentation on the MV networks of the ENGAGE dataset show similar performance results to those of this study.
    \item \textbf{Scheduled Sampling:} Though the effects of error accumulation were mild in our study, it is an inherent risk that comes with autoregressive sequence modeling. Thus, to mitigate error accumulation, scheduled sampling can be investigated, where the model is occasionally fed its own previous predictions during training. By gradually decreasing the probability of using ground-truth data, the model can develop self-correction mechanisms to mitigate the accumulation of recursive drifts along long feeder paths.
    \item \textbf{Latency and Parallelization:} Implementing the BFS traversal in a lower-level language like C++ and parallelizing independent feeder branches by tree depth could reduce the inference bottleneck to $O(\log N)$ latency, making the scaling more comparable to globally-predictive methods.
    \item \textbf{Alternative Architectures:} Exploring the use of other sequence models, such as LSTMs or Transformers, could potentially capture more complex long-range spatial dependencies within the radial paths. However, this would likely come at the cost of greatly increased training time, an additional   memory footprint, and lower interpretability.
    
\end{itemize}
\section{Conclusion}

This study introduces BOOST-RPF, a novel paradigm for power flow analysis that reformulates voltage prediction from a global graph-based regression task into a sequential path-based learning problem. By shifting the focus from joint nodal predictions to localized, recursive voltage-drop mappings, we aimed to reconcile the high performance of machine learning with the robust inductive biases of classical analytical solvers. The experimental results demonstrate that the BOOST-RPF framework, particularly the Parent-Residual variant, convincingly surpasses state-of-the-art neural baselines in both accuracy and robustness. By training on edge-level voltage drops rather than global snapshots, the model remains size-agnostic and demonstrates superior generalization to unseen network topologies (OOD) when compared to global Multi-Layer Perceptrons and Graph Neural Networks. Furthermore, the use of boosted tree ensembles allows for training times that are orders of magnitude shorter than neural networks (completing in seconds on a CPU) while achieving significantly higher sample efficiency through path-based learning. While the method exhibits a predictable $O(N)$ computational scaling, the absolute inference times remain well within the requirements for real-time monitoring and represent a manageable linear upper bound on latency.

Looking forward, the high generalization and speed of BOOST-RPF make it a promising candidate for the next generation of active distribution system management. Its ability to provide reliable and efficient power flow solutions is critical for real-time applications such as optimal voltage control, dynamic hosting capacity assessment, and the integration of highly volatile distributed energy resources. Future work will focus on extending this sequential paradigm to meshed topologies and implementing lower-level parallelization to further reduce inference latency for large-scale utility grids. To support further research and community adoption, we have released our models, experimental pipeline, and analysis scripts in an open-source repository\footnote{\url{https://github.com/EOkoyomon/boost-rpf}}, ensuring that our results are fully reproducible and the framework can be readily extended to other distribution-level applications.

\bibliographystyle{IEEEtranN}
\bibliography{references}

\newpage
\appendices
\section{DistFlow Formulation} \label{app:distflow}

Consider a radial distribution network represented by a directed graph $\mathcal{G} = (\mathcal{N}, \mathcal{E})$, where $\mathcal{N}$ is the set of buses and $\mathcal{E}$ is the set of lines. For each line $(i, j) \in \mathcal{E}$ connecting bus $i$ to bus $j$, let $z_{ij} = r_{ij} + \mathbf{j}x_{ij}$ denote the complex impedance. Let $V_i$ be the complex voltage magnitude at bus $i$, and let $S_{ij} = P_{ij} + \mathbf{j}Q_{ij}$ be the complex power flowing from bus $i$ to bus $j$.
The non-linear DistFlow equations \cite{baran1989network} describing the voltage drop and power balance are:
\begin{align}
    V_j^2 &= V_i^2 - 2(r_{ij}P_{ij} + x_{ij}Q_{ij}) + (r_{ij}^2 + x_{ij}^2)\frac{P_{ij}^2 + Q_{ij}^2}{V_i^2} \label{eq:distflow_v} \\
    P_{ij} &= P_j^{L} + \sum_{k: (j, k) \in \mathcal{E}} \left( P_{jk} + r_{jk}\frac{P_{jk}^2 + Q_{jk}^2}{V_j^2} \right) \label{eq:distflow_p} \\
    Q_{ij} &= Q_j^{L} + \sum_{k: (j, k) \in \mathcal{E}} \left( Q_{jk} + x_{jk}\frac{P_{jk}^2 + Q_{jk}^2}{V_j^2} \right) \label{eq:distflow_q}
\end{align}
where $P_j^L$ and $Q_j^L$ represent the active and reactive load consumption at node $j$.

\textit{Voltage Angle Recovery:}
While the original formulation focused on voltage magnitudes and losses, the voltage angles can be recovered using a complementary linearization often utilized in convex relaxations of Optimal Power Flow, as detailed by \citet{farivar2013branch}, and a similar set of common assumptions.
Starting from the voltage drop equation $V_j = V_i - (r_{ij} + \mathbf{j}x_{ij})I_{ij}$ and applying the small-angle approximation ($\sin(\theta_{ij}) \approx \theta_i - \theta_j$ and $\cos(\theta_{ij}) \approx 1$), the voltage angle difference is decoupled from the voltage magnitude. The resulting linear relationship is:
\begin{equation}
    \theta_j \approx \theta_i - \frac{x_{ij}P_{ij} - r_{ij}Q_{ij}}{V_{nom}^2}
    \label{eq:lindistflow_angle}
\end{equation}
where $\theta$ is the voltage angle in radians and $V_{nom}$ is the nominal voltage (typically approximated as the slack bus voltage magnitude).

\section{Model Architectures and Hyperparameters} \label{app:model-params}

To ensure a fair and rigorous comparison between our proposed BOOST-RPF variants and the established baselines, we performed extensive hyperparameter tuning for all architectures. The tuning procedure followed a two-stage approach: initially, a random search across 20 configurations was conducted to survey the parameter space, followed by a targeted grid search (approximately 8–12 configurations) to refine the best-performing trends.

\subsection{XGBoost Architecture}

The three BOOST-RPF variants (XGB-Absolute, XGB-Parent, and XGB-LDF) utilize the XGBoost regressor as their core sequential engine. These models were trained on CPU-based hardware. For all XGBoost models, we utilized the standard ``\texttt{reg:squarederror}'' objective function, which minimizes the mean squared error between the predicted and true voltage targets to ensure the regression process converges on the most accurate estimates.

\subsubsection{Hyperparameter Tuning Space}

The tuning script evaluated the following ranges. For all XGB models we used the standard 

\begin{itemize}
    \item \texttt{n\_estimators}: [50, 100, 200, 300, 400] - Controls the number of boosted trees (iterations) to prevent under- or over-fitting.
    \item \texttt{max\_depth}: [5, 7, 9, 10, 12] - Limits the maximum depth of each tree, controlling model complexity.
    \item \texttt{learning\_rate}: [0.1, 0.2, 0.5, 1.0] - Determines the step size shrinkage at each boosting iteration to improve robustness.
    \item \texttt{min\_child\_weight}: [1, 5, 10, 20] - Sets the minimum sum of instance weights needed in a child node to control conservative splitting.
    \item \texttt{subsample}: [0.8, 0.9, 1.0] — The fraction of training data sampled for each tree to prevent overfitting.
    \item \texttt{colsample\_bytree}: [0.8, 1.0] - The fraction of features sampled for each tree.
    \item \texttt{multi\_strategy}: [``multi\_output\_tree'', ``one\_output\_per\_tree''] - Determines whether a single tree predicts both magnitude and angle (multi-output) or separate trees are used for each.
\end{itemize}

\subsubsection{Final Configuration} 

For all XGBoost variants, we use the same base model configuration to maintain a fair comparison. Variants only differ in terms of their prediction target, as explained in Section \ref{method:sequential-learning-architecture}.

\begin{itemize}
    \item \texttt{n\_estimators}: 200
    \item \texttt{max\_depth}: 7
    \item \texttt{learning\_rate}: 0.5
    \item \texttt{min\_child\_weight}: 5
    \item \texttt{subsample}: 0.9
    \item \texttt{colsample\_bytree}: 1.0
    \item \texttt{multi\_strategy}: ``multi\_output\_tree''
\end{itemize}


\subsection{ARMA-GNN Architecture}

Since \citet{hansen_power_2022} released their ARMA-GNN open-source, we simply adapt their implementation to PyTorch from its original implementation in TensorFlow. We additionally correct the open source implementation to match the paper's description, and ultimately improving the model's speed and performance. We ensure that we receive comparable scores to the publication before proceeding with further fine tuning. Using our improved re-implementation, training speed decreased from 80mins to 15 mins, while the test NRMSE improved from 0.022 to 0.019 for the same-grid experiment with IEEE Case 30.

The ARMA-GNN is designed to capture graph signals across power networks. The inital configuration of the model consistent of two fully-connected pre-processing layers, five parallel stacks of 8-layer ARMA-GNN convolutions, and two post-processing layers. All of these layers had 64 hidden units, and use Leaky ReLU activations with a negative slope coefficient of $\alpha=0.2$. It was trained for 250 epochs using the ADAM optimizer with learning rate 0.001, batch size 16, and a mean squared error (MSE) loss. Where not explicitly mentioned, we keep the same architectural design choices.

\subsubsection{Hyperparameter Tuning Space}

The tuning script evaluated the following ranges. For all XGB models we used the standard 

\begin{itemize}
    \item \texttt{batch\_size}: [16, 32, 64] - The number of graph samples processed per training step, balancing gradient stability and stochasticity.
    \item \texttt{learning\_rate}: [0.001, 0.0001, 0.00001] - Controls the optimizer's step size, critical for ensuring stable convergence characteristics.
    \item \texttt{num\_layers}: [6, 7, 8, 9, 10] - The number of ARMA convolution layers in each stack, used to aggregate information from distant neighborhoods.
    \item \texttt{hidden\_dim}: [32, 64, 128, 256] - The width of the latent feature, which determines representation capacity.
    \item \texttt{dropout}: [0.0, 0.1, 0.2] - Regularization to prevent co-adaptation of weights, by randomly deactivating neurons.
\end{itemize}

\subsubsection{Final Configuration}

The final training procedure remains largely similar to the initial training pipeline from the \citet{hansen_power_2022}. However, while the original reference paper utilized standard MSE, our pipeline achieves superior performance using a Normalized MSE loss function. This approach weighs the loss by the inverse of the target vector’s norm, ensuring that voltage magnitude (p.u.) and voltage angle (degrees) exert equal influence on the gradient updates despite their differing numerical scales. The final model configuration is as follows.

\noindent\textbf{Architecture:} 2 pre-dense layers, 8 ARMA convolution layers (with 5 stacks each), 2 post-dense layers.

\noindent\textbf{Hyperparameters:}
\begin{itemize}
    \item \texttt{batch\_size}: 64
    \item \texttt{learning\_rate}: 0.001
    \item \texttt{hidden\_dim}: 64
    \item \texttt{dropout}: 0.0
    \item \texttt{loss\_function}: Normalized MSE
\end{itemize}

\noindent\textbf{Activations:} LeakyReLU with a negative slope of 0.2.


\subsection{Global MLP Architecture}

The GlobalMLP serves as a topology-agnostic baseline, processing all bus and edge features globally. To handle the heterogeneous nature of the datasets, it is configured to accommodate the largest grid size in the suite (129 nodes from LV Rural3). This is done by flattening all the node and edge features of the network into a single feature vector to learn from. The original baseline from \citet{hansen_power_2022} consists of bus and branch pre-processing layers with 64 units each. These are then concatenated and passing into two hidden layers with 128 units before mapping to the final output dimension. The processing layers and the hidden layers use Leaky ReLU activations with $\alpha=0.2$. Their GlobalMLP was trained for 250 epochs using the ADAM optimizer with learning rate 0.001, batch size 16, and an MSE loss. 

\subsubsection{Hyperparameter Tuning Space}

\begin{itemize}
    \item \texttt{batch\_size}: [16, 32, 64]
    \item \texttt{learning\_rate}: [0.001, 0.0001, 0.00001, 0.000001]
    \item \texttt{hidden\_dim}: [64, 128, 256, 512] - Controls the capacity of the core dense layers.
    \item \texttt{dropout}: [0.0, 0.1, 0.2]
\end{itemize}

\subsubsection{Final Configuration}

Similarly, we found that using the Normalized MSE loss provided more accurate and robust training procedures and final predictions when compared to the standard MSE loss. As in the reference paper, we opt to keep the network depth fixed at two core layers to prevent the parameter count from ballooning beyond other architectures and risking too much overfitting.

\noindent\textbf{Architecture:} Bus and edge pre-processing layers followed by two core dense layers.

\noindent\textbf{Hyperparameters:}
\begin{itemize}
    \item \texttt{batch\_size}: 64
    \item \texttt{learning\_rate}: 0.000001
    \item \texttt{hidden\_dim}: 256
    \item \texttt{dropout}: 0.0
    \item \texttt{loss\_function}: Normalized MSE
\end{itemize}

\noindent\textbf{Activations:} LeakyReLU with a negative slope of 0.2.

\section{Raw Test Performance Results}

\begin{table*}[htbp]
\centering
\caption{Model Performance Metrics by Testing Grid}
\label{tab:raw_model_performance}
\begin{tabular}{llccccr}
\toprule
\textbf{Testing Grid} & \textbf{Model} & \textbf{RMSE VM} & \textbf{RMSE VA} & \textbf{Train (s)} & \textbf{Inference (ms)} \\
\midrule
Kerber Dorfnetz & GlobalMLP & 0.00680 {\scriptsize $\pm$ 0.00024} & 0.18232 {\scriptsize $\pm$ 0.00328} & 336.3 {\scriptsize $\pm$ 6.0} & \textbf{0.31797 {\scriptsize $\pm$ 0.00742}} \\
Kerber Dorfnetz & ARMA-GNN & 0.01045 {\scriptsize $\pm$ 0.00010} & 0.37849 {\scriptsize $\pm$ 0.18130} & 986.5 {\scriptsize $\pm$ 170.9} & 2.26350 {\scriptsize $\pm$ 0.18526} \\
Kerber Dorfnetz & XGB-Absolute & 0.00413 {\scriptsize $\pm$ 0.00012} & 0.02071 {\scriptsize $\pm$ 0.00126} & 10.4 {\scriptsize $\pm$ 0.5} & 9.10650 {\scriptsize $\pm$ 0.52861} \\
Kerber Dorfnetz & XGB-Parent & \textbf{0.00084 {\scriptsize $\pm$ 0.00003}} & \textbf{0.00734 {\scriptsize $\pm$ 0.00025}} & 11.7 {\scriptsize $\pm$ 0.4} & 9.20448 {\scriptsize $\pm$ 0.49567} \\
Kerber Dorfnetz & XGB-LDF & 0.00253 {\scriptsize $\pm$ 0.00004} & 0.01117 {\scriptsize $\pm$ 0.00009} & 11.4 {\scriptsize $\pm$ 0.4} & 9.16345 {\scriptsize $\pm$ 0.65104} \\
Kerber Dorfnetz & DistFlow & 0.03422 {\scriptsize $\pm$ 0.00039} & 0.43964 {\scriptsize $\pm$ 0.00044} & \textbf{0.0 {\scriptsize $\pm$ 0.0}} & 2.67949 {\scriptsize $\pm$ 0.01238} \\
\midrule
All (Known) & GlobalMLP & 0.00559 {\scriptsize $\pm$ 0.00018} & 0.14546 {\scriptsize $\pm$ 0.00519} & 174.9 {\scriptsize $\pm$ 20.3} & \textbf{0.33037 {\scriptsize $\pm$ 0.02253}} \\
All (Known) & ARMA-GNN & 0.04209 {\scriptsize $\pm$ 0.00904} & 0.36551 {\scriptsize $\pm$ 0.26993} & 1095.5 {\scriptsize $\pm$ 92.1} & 2.23788 {\scriptsize $\pm$ 0.17569} \\
All (Known) & XGB-Absolute & 0.00489 {\scriptsize $\pm$ 0.00011} & 0.02672 {\scriptsize $\pm$ 0.00229} & 9.9 {\scriptsize $\pm$ 0.7} & 5.76496 {\scriptsize $\pm$ 0.53010} \\
All (Known) & XGB-Parent & \textbf{0.00073 {\scriptsize $\pm$ 0.00005}} & \textbf{0.01123 {\scriptsize $\pm$ 0.00271}} & 10.8 {\scriptsize $\pm$ 0.8} & 5.91721 {\scriptsize $\pm$ 0.68147} \\
All (Known) & XGB-LDF & 0.00518 {\scriptsize $\pm$ 0.00022} & 0.04775 {\scriptsize $\pm$ 0.00301} & 11.1 {\scriptsize $\pm$ 1.0} & 6.25072 {\scriptsize $\pm$ 0.76570} \\
All (Known) & DistFlow & 0.02828 {\scriptsize $\pm$ 0.00012} & 1.09591 {\scriptsize $\pm$ 0.12172} & \textbf{0.0 {\scriptsize $\pm$ 0.0}} & 1.86729 {\scriptsize $\pm$ 0.01424} \\
\midrule
LV-rural1 & GlobalMLP & 2.18025 {\scriptsize $\pm$ 0.41166} & 93.33799 {\scriptsize $\pm$ 3.23029} & 218.1 {\scriptsize $\pm$ 54.0} & \textbf{0.29875 {\scriptsize $\pm$ 0.03797}} \\
LV-rural1 & ARMA-GNN & 0.01096 {\scriptsize $\pm$ 0.00207} & 1.93481 {\scriptsize $\pm$ 0.02017} & 1397.9 {\scriptsize $\pm$ 273.7} & 2.18000 {\scriptsize $\pm$ 0.16065} \\
LV-rural1 & XGB-Absolute & 0.01591 {\scriptsize $\pm$ 0.00317} & 1.77800 {\scriptsize $\pm$ 0.14991} & 13.0 {\scriptsize $\pm$ 1.3} & 1.11777 {\scriptsize $\pm$ 0.13147} \\
LV-rural1 & XGB-Parent & \textbf{0.00992 {\scriptsize $\pm$ 0.00074}} & \textbf{1.62739 {\scriptsize $\pm$ 0.01958}} & 13.7 {\scriptsize $\pm$ 1.0} & 1.60736 {\scriptsize $\pm$ 0.82335} \\
LV-rural1 & XGB-LDF & 0.01358 {\scriptsize $\pm$ 0.00423} & 3.34088 {\scriptsize $\pm$ 0.09296} & 13.2 {\scriptsize $\pm$ 1.1} & 1.78556 {\scriptsize $\pm$ 0.52970} \\
LV-rural1 & DistFlow & 0.03036 {\scriptsize $\pm$ 0.00000} & 4.34720 {\scriptsize $\pm$ 0.00000} & \textbf{0.0 {\scriptsize $\pm$ 0.0}} & 0.48836 {\scriptsize $\pm$ 0.01197} \\
\midrule
LV-rural2 & GlobalMLP & 2.12559 {\scriptsize $\pm$ 0.57411} & 57.97328 {\scriptsize $\pm$ 5.76042} & 170.2 {\scriptsize $\pm$ 28.5} & \textbf{0.30125 {\scriptsize $\pm$ 0.03721}} \\
LV-rural2 & ARMA-GNN & 0.05981 {\scriptsize $\pm$ 0.03709} & 7.80871 {\scriptsize $\pm$ 6.53410} & 1029.8 {\scriptsize $\pm$ 458.7} & 2.38566 {\scriptsize $\pm$ 0.23235} \\
LV-rural2 & XGB-Absolute & 0.02408 {\scriptsize $\pm$ 0.00045} & 0.14206 {\scriptsize $\pm$ 0.03222} & 9.7 {\scriptsize $\pm$ 0.8} & 7.47414 {\scriptsize $\pm$ 0.54942} \\
LV-rural2 & XGB-Parent & 0.01684 {\scriptsize $\pm$ 0.00076} & 0.05168 {\scriptsize $\pm$ 0.00308} & 10.3 {\scriptsize $\pm$ 0.8} & 7.60346 {\scriptsize $\pm$ 0.31866} \\
LV-rural2 & XGB-LDF & 0.02293 {\scriptsize $\pm$ 0.00136} & 0.29370 {\scriptsize $\pm$ 0.15444} & 10.4 {\scriptsize $\pm$ 0.8} & 7.68416 {\scriptsize $\pm$ 0.47870} \\
LV-rural2 & DistFlow & \textbf{0.01400 {\scriptsize $\pm$ 0.00000}} & \textbf{0.04265 {\scriptsize $\pm$ 0.00000}} & \textbf{0.0 {\scriptsize $\pm$ 0.0}} & 2.32253 {\scriptsize $\pm$ 0.04741} \\
\midrule
LV-rural3 & GlobalMLP & 1.12972 {\scriptsize $\pm$ 0.13843} & 67.85296 {\scriptsize $\pm$ 0.83434} & 177.8 {\scriptsize $\pm$ 33.5} & \textbf{0.29697 {\scriptsize $\pm$ 0.03876}} \\
LV-rural3 & ARMA-GNN & 0.02259 {\scriptsize $\pm$ 0.00353} & 0.94820 {\scriptsize $\pm$ 0.30974} & 1640.3 {\scriptsize $\pm$ 385.1} & 2.32118 {\scriptsize $\pm$ 0.15272} \\
LV-rural3 & XGB-Absolute & 0.02666 {\scriptsize $\pm$ 0.00417} & 0.85377 {\scriptsize $\pm$ 0.06810} & 10.5 {\scriptsize $\pm$ 1.2} & 10.08740 {\scriptsize $\pm$ 0.41160} \\
LV-rural3 & XGB-Parent & \textbf{0.01818 {\scriptsize $\pm$ 0.00332}} & 1.07782 {\scriptsize $\pm$ 0.40653} & 10.8 {\scriptsize $\pm$ 0.7} & 9.84262 {\scriptsize $\pm$ 0.62003} \\
LV-rural3 & XGB-LDF & 0.01873 {\scriptsize $\pm$ 0.00078} & \textbf{0.46609 {\scriptsize $\pm$ 0.01594}} & 10.7 {\scriptsize $\pm$ 0.6} & 9.99387 {\scriptsize $\pm$ 0.59400} \\
LV-rural3 & DistFlow & 0.03071 {\scriptsize $\pm$ 0.00000} & 1.03803 {\scriptsize $\pm$ 0.00000} & \textbf{0.0 {\scriptsize $\pm$ 0.0}} & 3.02180 {\scriptsize $\pm$ 0.02302} \\
\midrule
LV-semiurb4 & GlobalMLP & 1.19228 {\scriptsize $\pm$ 0.11436} & 52.57844 {\scriptsize $\pm$ 1.48481} & 234.3 {\scriptsize $\pm$ 68.9} & \textbf{0.29400 {\scriptsize $\pm$ 0.03582}} \\
LV-semiurb4 & ARMA-GNN & 0.03648 {\scriptsize $\pm$ 0.02040} & 0.37253 {\scriptsize $\pm$ 0.03609} & 2262.6 {\scriptsize $\pm$ 606.0} & 2.56294 {\scriptsize $\pm$ 0.75073} \\
LV-semiurb4 & XGB-Absolute & \textbf{0.01320 {\scriptsize $\pm$ 0.00268}} & 0.55668 {\scriptsize $\pm$ 0.10366} & 13.0 {\scriptsize $\pm$ 1.0} & 3.38661 {\scriptsize $\pm$ 0.25575} \\
LV-semiurb4 & XGB-Parent & 0.01595 {\scriptsize $\pm$ 0.00742} & 0.12819 {\scriptsize $\pm$ 0.05212} & 13.2 {\scriptsize $\pm$ 0.7} & 3.46267 {\scriptsize $\pm$ 0.18223} \\
LV-semiurb4 & XGB-LDF & 0.02293 {\scriptsize $\pm$ 0.00167} & 0.61619 {\scriptsize $\pm$ 0.01431} & 13.4 {\scriptsize $\pm$ 0.9} & 3.38982 {\scriptsize $\pm$ 0.27562} \\
LV-semiurb4 & DistFlow & 0.02395 {\scriptsize $\pm$ 0.00000} & \textbf{0.10019 {\scriptsize $\pm$ 0.00000}} & \textbf{0.0 {\scriptsize $\pm$ 0.0}} & 1.13786 {\scriptsize $\pm$ 0.03385} \\
\midrule
LV-semiurb5 & GlobalMLP & 1.56605 {\scriptsize $\pm$ 0.29778} & 43.45174 {\scriptsize $\pm$ 1.32058} & 178.1 {\scriptsize $\pm$ 19.9} & \textbf{0.29953 {\scriptsize $\pm$ 0.05459}} \\
LV-semiurb5 & ARMA-GNN & 0.01874 {\scriptsize $\pm$ 0.00729} & 0.78177 {\scriptsize $\pm$ 0.16304} & 1539.2 {\scriptsize $\pm$ 550.8} & 2.78184 {\scriptsize $\pm$ 1.12949} \\
LV-semiurb5 & XGB-Absolute & 0.02239 {\scriptsize $\pm$ 0.00189} & \textbf{0.32724 {\scriptsize $\pm$ 0.17925}} & 10.4 {\scriptsize $\pm$ 0.9} & 8.68968 {\scriptsize $\pm$ 0.65682} \\
LV-semiurb5 & XGB-Parent & \textbf{0.00761 {\scriptsize $\pm$ 0.00157}} & 1.57089 {\scriptsize $\pm$ 0.02248} & 10.4 {\scriptsize $\pm$ 0.5} & 8.62050 {\scriptsize $\pm$ 0.52529} \\
LV-semiurb5 & XGB-LDF & 0.01559 {\scriptsize $\pm$ 0.00292} & 0.33221 {\scriptsize $\pm$ 0.09795} & 10.7 {\scriptsize $\pm$ 0.8} & 8.68134 {\scriptsize $\pm$ 0.66768} \\
LV-semiurb5 & DistFlow & 0.03633 {\scriptsize $\pm$ 0.00000} & 0.85845 {\scriptsize $\pm$ 0.00000} & \textbf{0.0 {\scriptsize $\pm$ 0.0}} & 2.68880 {\scriptsize $\pm$ 0.03093} \\
\midrule
LV-urban6 & GlobalMLP & 0.72959 {\scriptsize $\pm$ 0.08604} & 71.87856 {\scriptsize $\pm$ 0.70576} & 239.2 {\scriptsize $\pm$ 92.6} & \textbf{0.30419 {\scriptsize $\pm$ 0.02921}} \\
LV-urban6 & ARMA-GNN & 0.02666 {\scriptsize $\pm$ 0.01905} & 0.72735 {\scriptsize $\pm$ 0.18299} & 1414.1 {\scriptsize $\pm$ 310.3} & 2.71810 {\scriptsize $\pm$ 1.02225} \\
LV-urban6 & XGB-Absolute & 0.01766 {\scriptsize $\pm$ 0.00084} & 0.20312 {\scriptsize $\pm$ 0.05062} & 13.2 {\scriptsize $\pm$ 0.9} & 4.61742 {\scriptsize $\pm$ 0.35008} \\
LV-urban6 & XGB-Parent & \textbf{0.01218 {\scriptsize $\pm$ 0.00296}} & 0.14993 {\scriptsize $\pm$ 0.01861} & 13.3 {\scriptsize $\pm$ 1.0} & 4.75147 {\scriptsize $\pm$ 0.35644} \\
LV-urban6 & XGB-LDF & 0.01829 {\scriptsize $\pm$ 0.00244} & 0.78574 {\scriptsize $\pm$ 0.01269} & 13.3 {\scriptsize $\pm$ 0.5} & 4.66907 {\scriptsize $\pm$ 0.37751} \\
LV-urban6 & DistFlow & 0.03030 {\scriptsize $\pm$ 0.00000} & \textbf{0.11978 {\scriptsize $\pm$ 0.00000}} & \textbf{0.0 {\scriptsize $\pm$ 0.0}} & 1.50890 {\scriptsize $\pm$ 0.02358} \\
\bottomrule
\end{tabular}
\end{table*}

\end{document}